\documentclass{article}
\usepackage{amsmath,graphicx,amsfonts}

\def\x{{\mathbf X}}
\def\xhat{\hat{\mathbf X}}
\def\c{{\mathbf C}}
\def\cbar{{\bar{\mathbf C}}}

\title{Denoising without access to clean data \\ using a partitioned autoencoder}

\author
  {Dan Stowell\\
	Centre for Digital Music\\
	School of Elec. Eng. and Computer Science\\
	Queen Mary University of London
\and
	Richard E. Turner\\
	Computational and Biological Learning Lab\\
	Department of Engineering\\
	University of Cambridge}

\begin{document}
\maketitle
\begin{abstract}
Training a \textit{denoising autoencoder} neural network requires access to truly clean data, a requirement which is often im\-practical.
To remedy this, we introduce a method to train an autoencoder using only noisy data, having examples with and without the signal class of interest.
The autoencoder learns a partitioned representation of signal and noise, learning to reconstruct each separately.
We illustrate the method by denoising birdsong audio (available abundantly in uncontrolled noisy datasets) using a convolutional autoencoder.
\end{abstract}
%
\section{Introduction}
\label{sec:intro}

An autoencoder (AE) is a neural network trained in unsupervised fashion,
to encode its input to some latent representation
and to decode that representation to a faithful reconstruction of its input.
The autoencoder can then be used as a codec,
or to convert data to its latent representation for downstream processing such as classification.
The \textit{denoising autoencoder} (DAE) is a variant of this in which the inputs are combined with some corruption (such as additive noise or masking),
and the system is trained to recover the clean, de-noised data \cite{Vincent:2010}.
The DAE training scheme can be used in denoising applications,
and is also a popular way to encourage the autoencoder to learn a more meaningful latent representation of the data.
Autoencoders including the DAE have yielded leading results in recent years in deep learning for signal processing \cite{Vincent:2010,Lecun:2015}.

However, there is a significant problem with the DAE approach which hampers its use in practical applications:
it may often be impossible to supply truly clean data.
This is common in our application example---natural sound recordings---but also for video, image and audio applications across many domains.
In fact, objects/events are often sparsely represented in data while background and other noise are densely represented,
meaning that it is often easy to provide ``noise-only'' examples while difficult to provide ``noise-free'' examples.

In this paper we propose an alternative approach to train an AE so that it can perform denoising,
given training data which can only be weakly labelled as ``noise-only'' or ``noise and possible signal''.
The system learns a partitioned latent representation,
with identifiable noise and signal coefficients,
and can then perform denoising and/or recover a signal-only latent representation for further analysis.
The method is general-purpose; we illustrate it here with an application to denoising birdsong audio spectrograms.

\section{Partitioned autoencoders}
\label{sec:method}

A standard AE learns a function of the form
$
 \xhat = g(f(\x))
$
where $\x$ is an input datum (a matrix, in this paper), and the autoencoder is composed of encoder $f(\cdot)$ and decoder $g(\cdot)$.
The DAE learns a function 
$
 \xhat = g(f(u(\x)))
$
where $u(\cdot)$ is a stochastic noise corruption process.
The training objective is such that $\xhat$ is encouraged to be as close to $\x$ as possible,
often $\sum_i{\|\x_i - \xhat_i\|^2}$ where $||\cdot||$ is the Frobenius norm and the sum is taken over a minibatch of training data.
(A minibatch is a small subset of training data used for one iteration of stochastic gradient descent [SGD].)
Given this objective, in practice a DAE will learn to denoise its input to the extent that $\x$ itself is clean,
having no incentive to ``overshoot'' and remove any noise that may be intrinsic to the original $\x$.
The latent representation output from $f(\cdot)$ parametrises the manifold 
on which the reconstructed signal data lie.
Information about the noise present in each datum is not captured,
except implicitly as $\x - \xhat$ if $\xhat$ is a good estimate.

Many equivalent parametrisations of the manifold may be possible,
some allowing a more semantic interpretation of each latent coefficient than others,
and the standard AE or DAE does not distinguish among these parametrisations.
There has been recent interest in adapting the training schemes of autoencoders such that the latent representation is explicitly semantic,
capturing attributes of the input in specific subsets of the latent variables \cite{Kulkarni:2015,Cheung:2015}.
We refer to these as \textit{partitioned autoencoders} since the latent variables are partitioned into subsets which are treated differently from each other during training.
Crucially, in this prior work the training scheme relies heavily on the existence of large structured datasets:
in \cite{Cheung:2015} a balanced dataset of labelled digit images;
in \cite{Kulkarni:2015} a dataset of faces constructed through systematic variation of attributes such as pose and lighting.
Without such known structure in the training data, their proposed training schemes will be either impossible to apply,
or biased by the presence of unbalanced or correlated factors in the training data.

Our present motivation is to learn a denoising representation, trained using data which does not contain truly clean examples.
If we can develop a scheme that learns to represent both signal and noise, but partitioning them into separate latent coefficients,
then we will be able to perform denoising or further analysis by using only the ``foreground'' (signal) coefficients
and setting the remainder to zero.
The scheme of \cite{Kulkarni:2015} would be appropriate if the SNR of each training example were known and systematically varied,
but in uncontrolled datasets this information is rarely available.
Instead we propose a scheme based on the observation that many scenarios consist of sparsely-present foreground and densely-present background,
and so noise-only training data is much easier to come by than signal-only.

\begin{figure}[t]
  \centering
  \centerline{\includegraphics[width=0.8\columnwidth,clip,trim=0mm 0mm 0mm 0mm]{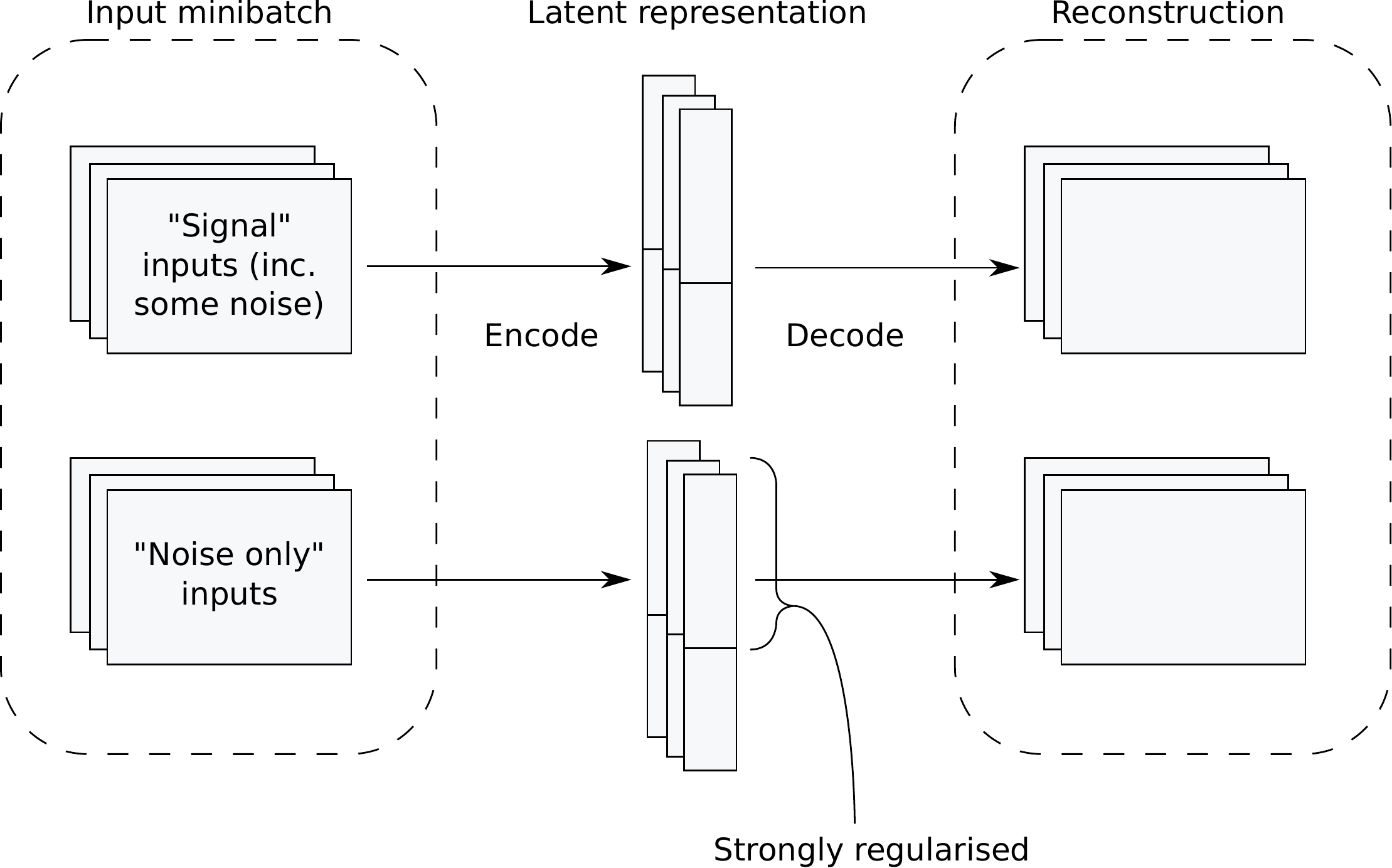}}
  \caption{Our training scheme aims to reconstruct all the minibatch items, while regularising some of the latents.}
  \label{fig:partitionaesystem_training}
\end{figure}

\begin{figure}[t]
  \centering
  \centerline{\includegraphics[width=0.6\columnwidth,clip,trim=0mm 0mm 0mm 0mm]{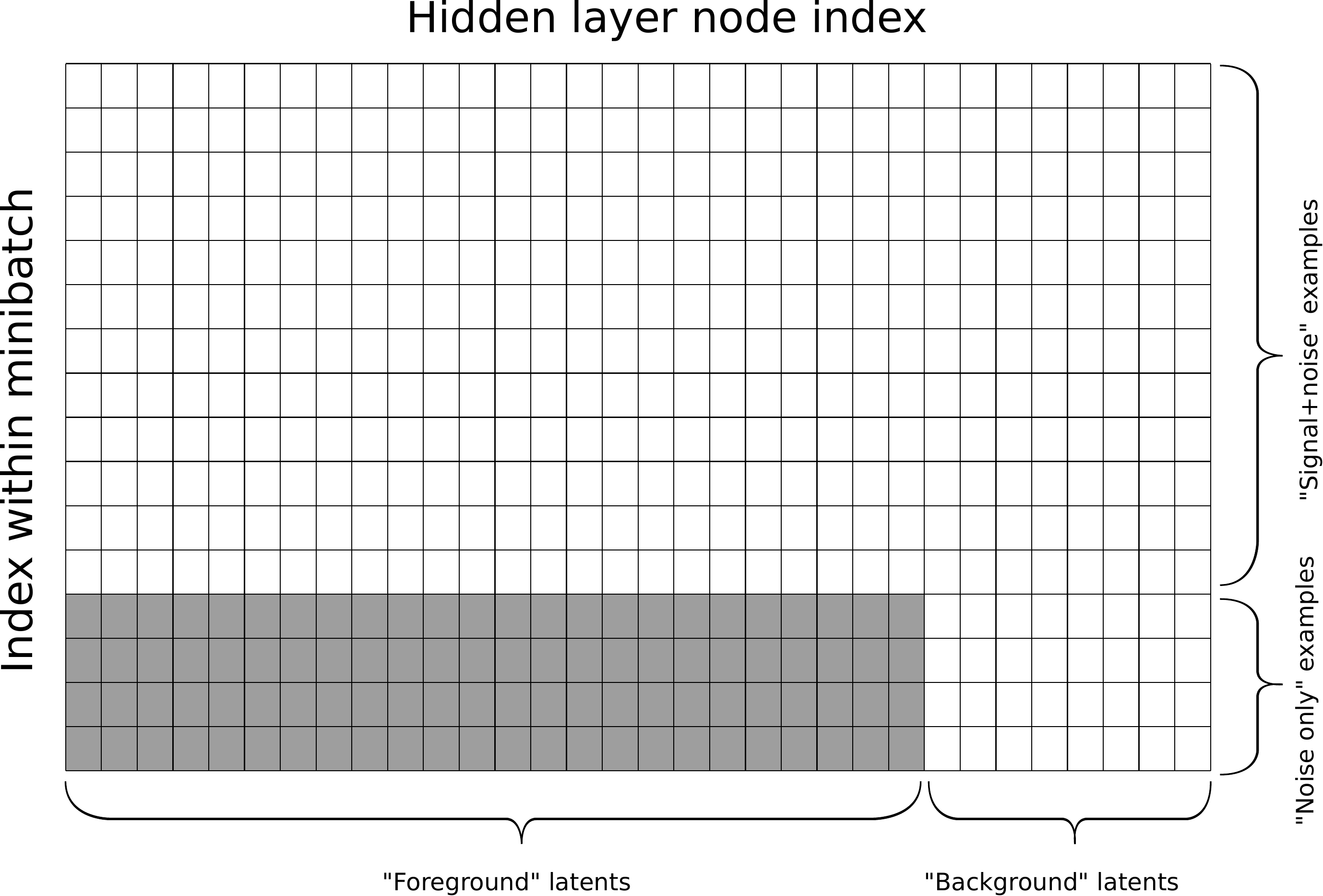}}
  \caption{Regularisation is applied to the subset of latents intended to represent signal, and only for noise-only examples.}
  \label{fig:partaegrid1}
\end{figure}

\begin{figure}[t]
  \centering
  \centerline{\includegraphics[width=0.8\columnwidth,clip,trim=0mm 0mm 0mm 0mm]{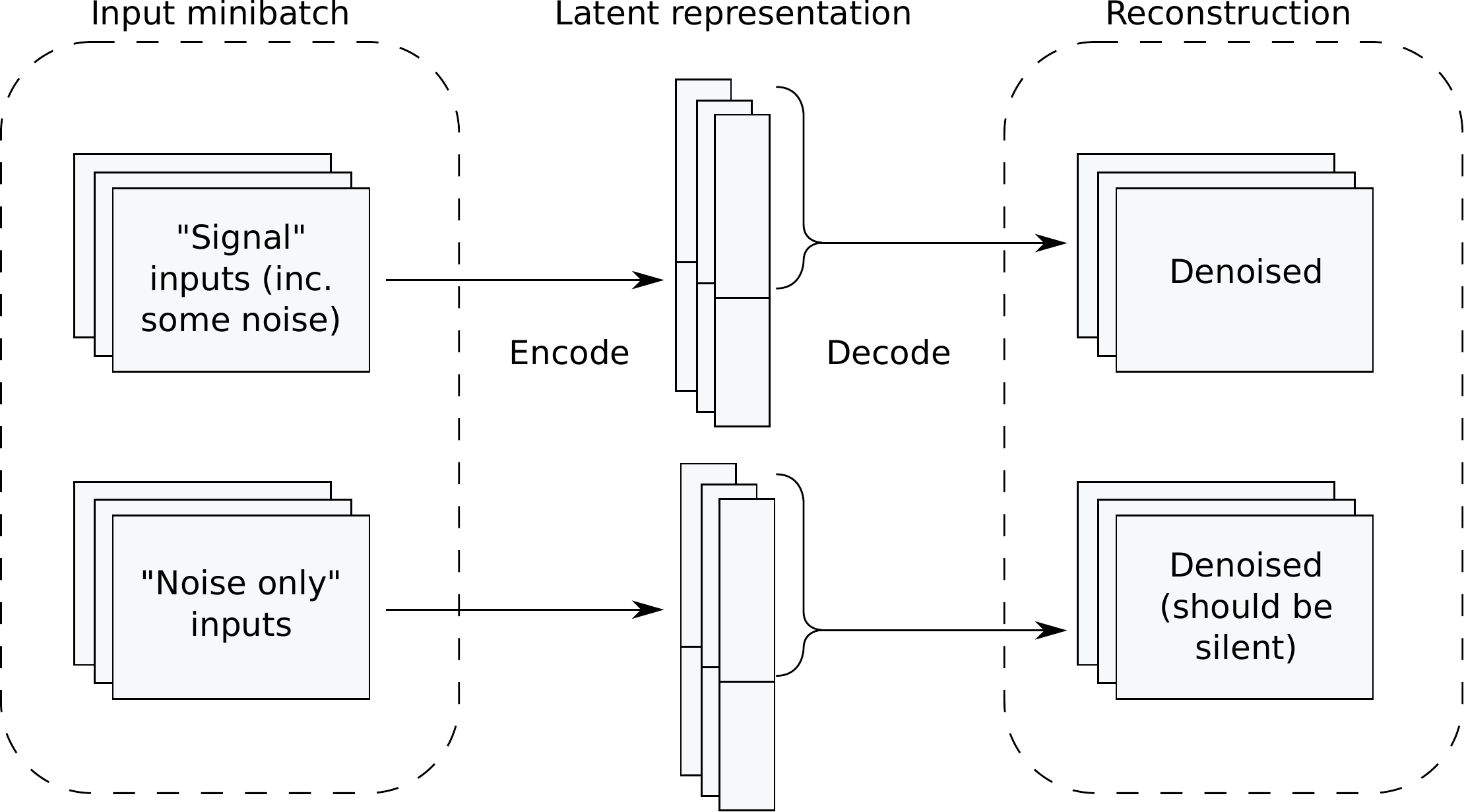}}
  \caption{Once trained, the system can denoise by reconstructing using only a subset of the latents (setting others to zero).}
  \label{fig:partitionaesystem_denoi}
\end{figure}

Our training scheme is based on a standard autoencoder reconstruction objective,
augmented with a structured regularisation of the latent variables.
In order to encourage the model to use the background latents to represent noise and never to represent signal,
we add a soft regulariser that penalises foreground latent activation for the noise-only examples.
For each training example $\x$ associated with a weak label $y$ taking the value 1 if the example is a ``noise-only'' example and 0 otherwise,
we train an autoencoder by minimising the following loss function:
$$
  l(\x, y) = \|\x - \xhat\|^2 + \frac{\lambda y}{\cbar} \|\c \odot f(\x)\|^2
$$
where
$\xhat = g(f(\x))$,
$\lambda$ is a regularisation coefficient,
$\odot$ represents elementwise multiplication,
$\c$ is a masking matrix containing 1 for latents which should represent foreground and 0 otherwise,
and $\cbar$ the mean value of $\c$.
We will construct our training minibatches with a fixed proportion of noise-only items at each iteration.
The value used for $\lambda$ will be relatively large, to impose a soft constraint pushing a subset of the latent values to zero in the case of noise-only items (Figure \ref{fig:partitionaesystem_training}, Figure \ref{fig:partaegrid1}).
This encourages the learned representation to use the non-regularised latents for the foreground signal.
The parameters of $f(\cdot)$ and $g(\cdot)$ will be optimised through SGD.
Once trained, denoising is achieved by reconstructing using \textit{only} the foreground latents, i.e.~setting the others to zero (Figure \ref{fig:partitionaesystem_denoi}).

Figure \ref{fig:partaegrid1} emphasises that the proportion of latents dedicated to foreground vs.\ background,
and the balance of ``signal-plus-noise'' and ``noise-only'' examples in a minibatch,
are independent configuration choices.
We tend to reserve 25\% of latents for background, and 25\% of a minibatch as noise-only;
in our evaluation we will evaluate the impact of varying the balance of latents.

Note that the regularisation scheme is asymmetric: it encourages some latents to zero in certain cases,
but it does not prevent them from being zero in the other cases.
In principle the system could simply never use those latents;
however the reconstruction cost encourages the system to use them to improve reconstruction in cases where it has that freedom.
The asymmetry also means that the important aspect of data labelling is to identify ``noise-only'' items with high precision.
If some ``noise-only'' items are not labelled as such, the system is free to represent them without making use of those latents.
The scheme can thus be used on datasets which are too large to label completely, but in which a subset of noise-only examples can be identified.

\section{Convolutional partitioned autoencoder for audio spectrograms}
\label{sec:convae}

The training scheme can be applied to various autoencoder architectures.
Convolutional neural networks have recently proven to be powerful for many tasks,
while relatively easy to train because they have far fewer free parameters than an equivalent
fully-connected network \cite{Lecun:2015,Dieleman:2014}.
In our application we aim to extract information from audio spectrograms, indexed by time and frequency.
We will thus use an autoencoder which is convolutional in time and fully-connected in frequency,
as is standard for recent neural network audio analysis \cite{Dieleman:2014}.
Given an input matrix $\x$ indexed by discrete time $n$ and frequency $h$,
we define our encoding function to be
$$
f(\x) = mp ( r ( {\mathbf W^c} \star (\x - \hat{\mu})/\hat{\sigma} ) )
$$
where
${\mathbf W^c}$ is a tensor of coding weights indexed by time $m$ frequency $h$ and latent index $k$,
$r(\cdot)$ is a rectified linear unit nonlinearity,
$mp(\cdot)$ represents the max-pooling operation applied along the time axis,
$\hat{\mu}$ and $\hat{\sigma}$ are the frequency-wise means and standard deviations estimated from the training set
and used for normalisation,
and $\star$ indicates one-dimensional convolution as follows:
$$
(A \star B)_{n,k} = \sum_{m=1}^M{\sum_{h=1}^H{ a_{m,h,k} b_{n-m,h} }} \quad n \in [1,N], k \in [1,K]
$$
resulting in a matrix indexed by time $n$ and latent index $k$.

Our decoding function is
$$
g(f(\x)) = {\mathbf W^d} \star mp^{-1}(f(\x))
$$
where
${\mathbf W^d}$ is a tensor of decoding weights,
and $mp^{-1}(\cdot)$ is the inverse of the max-pooling operation (the approximate inverse, as the non-maximal values are reconstituted by zeros).

In this temporally convolutional architecture, the mask matrix $\c$ is implemented as identical frame-wise mask matrices,
with $k$ indexing the set of latent time series.

Our data will be non-negative.
Data normalisation at the input is important for effective training, hence the use of $\hat{\mu}$ and $\hat{\sigma}$ \cite{Montavon:2012}.
However we do not undo the normalisation at the decoder outputs:
the non-negative target and the rectifier then give the property that the decoder learns a non-negative parts-based reconstruction.
We do not use bias units, as we found the resulting system easier to train (cf.\ \cite{Paine:2014,Memisevic:2014}).

For the evaluation that follows,
fixed configuration details are:
input spectrograms have 512 time frames and 32 frequency bins,
and we use 32 latent variables.
Convolution filters have a length of 9 time frames.
Our max-pooling downsamples the time axis by a factor of 16.
We train the network using AdaDelta to control the SGD learning rates \cite{Zeiler:2012}.
We do not use dropout.
We initialise the tensor of filters as a set of $K$ random orthogonal unit vectors of length $MH$, reshaping this to the tensor of shape $M \times H \times K$ (cf.\ \cite{Saxe:2013}).

In this study we explore only a single-layer autoencoder.
Our approach applies straightforwardly to a deep autoencoder,
and a deeper system would be expected to have a broader ability to generalise.


\begin{figure}[t]
  \centering
  \centerline{\includegraphics[width=0.55\columnwidth,clip,trim=20mm 81.5mm 20mm 15mm]{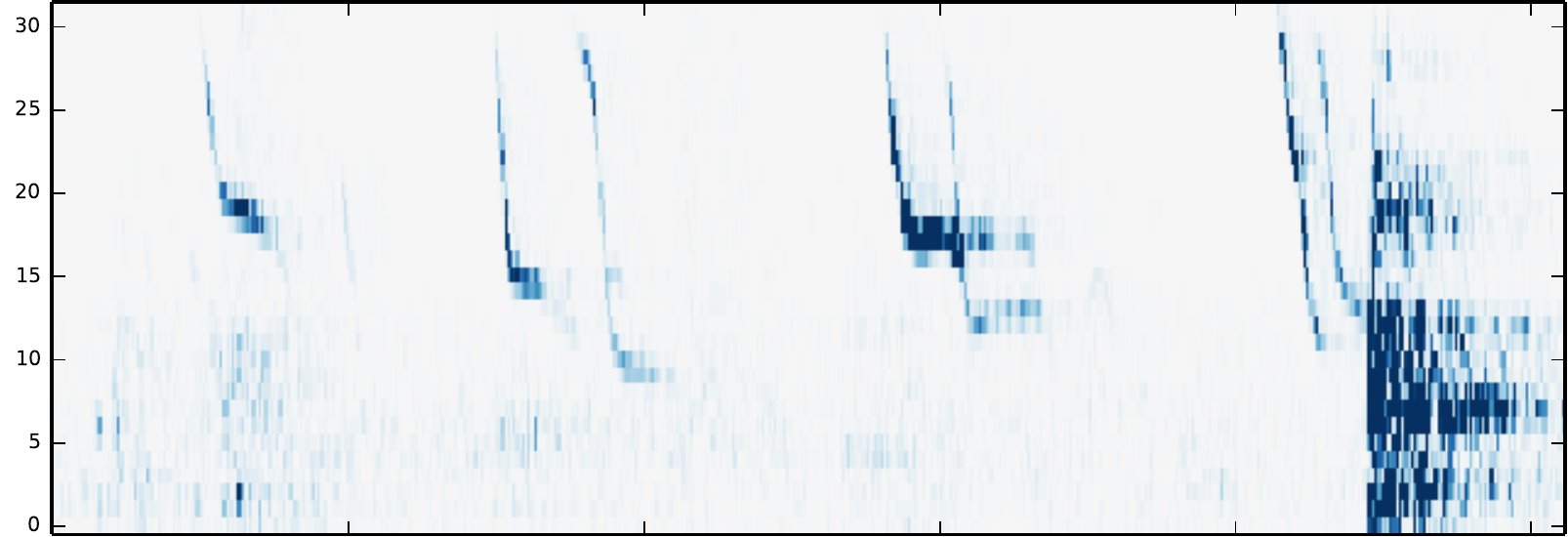}}
  \centerline{\includegraphics[width=0.55\columnwidth,clip,trim=20mm 81.5mm 20mm 15mm]{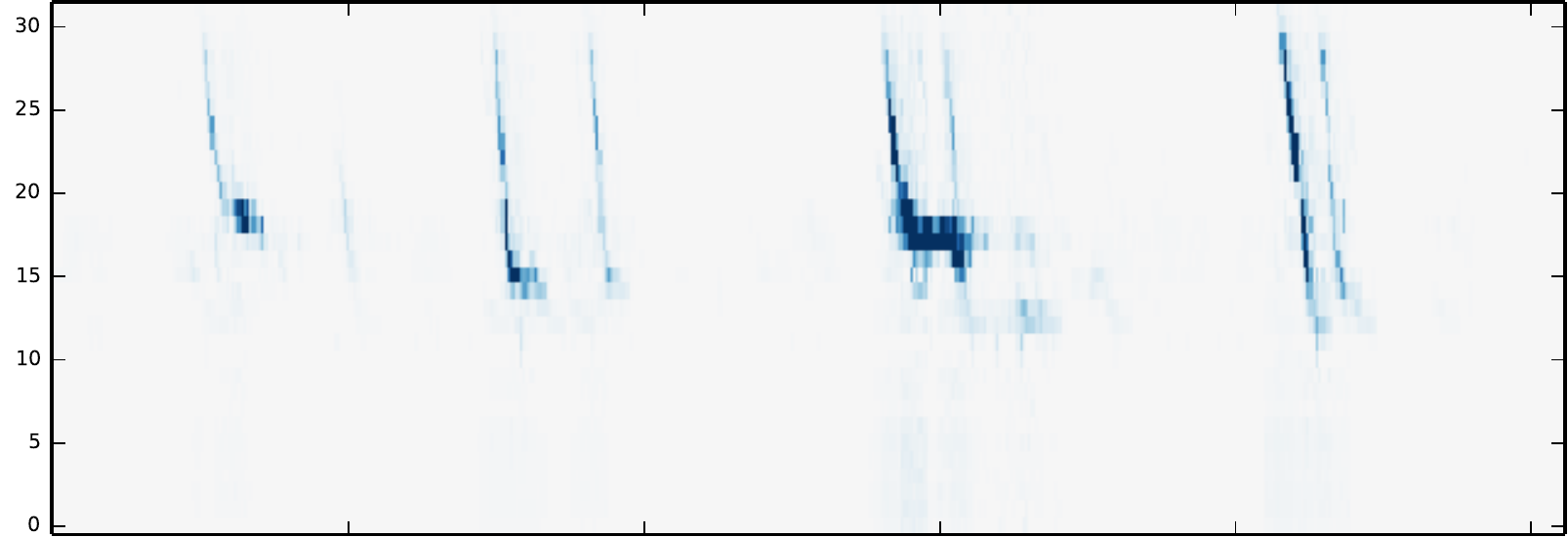}}
  \centerline{\includegraphics[width=0.55\columnwidth,clip,trim=20mm 15mm 20mm 81.5mm]{pics/cae1_feat_20150609_1227_1layer_inittedfromrandom_relu_2m_GREAT_training_p9}}
  \caption{Example of source separation with the proposed system. Upper: input spectrogram, 1.5 seconds (birdsong, plus background noise including a notable loud noise near the end). Middle: reconstructed signal spectrogram. Lower: reconstructed noise spectrogram.}
  \label{fig:spectrogramexample}
\end{figure}

\section{Evaluation}
\label{sec:eval}

We test our approach using a task to denoise birdsong audio spectrograms (Figure \ref{fig:spectrogramexample}).
As foreground we use recordings of chiff chaff (similarly to \cite{Stowell:2013,Stowell:2013d}).
For evaluation purposes we wish to add a background that offers a substantial test:
it should be diverse, nonstationary and contain significant energy in the same frequency band as the birdsong.
After considering many available options we settled on recorded restaurant noise,
which contains multi-speaker human speech as well as diverse percussive sound events.
We add this background noise, both to create a known amount of ``intrinsic'' noise in the examples,
and to create ``noise-only'' examples.

Our system is implemented in the Theano framework \cite{Theano:2012} making use of GPU processing.
To create a diverse training/validation dataset within the constraints of limited GPU memory,
we take advantage of the approximate additive nature of audio spectrograms as follows.
In each experiment we load 30 seconds of signal, of intrinsic noise and extrinsic noise,
and store their spectrograms to the GPU.
Then to generate each ``signal-plus-noise'' datum we randomly sample 1.5 second segments from each of the three sources and mix them.
To generate a ``noise only'' datum we sample only from the extrinsic noise source.
This creates a large generative dataset within limited memory.
We also experimented with replacing the 30 second source material regularly throughout training but this made little difference,
so we do not employ that in the present results.

\begin{figure}[t]
  \centering
  \centerline{\includegraphics[width=0.99\columnwidth,clip,trim=0mm 0mm 0mm 0mm]{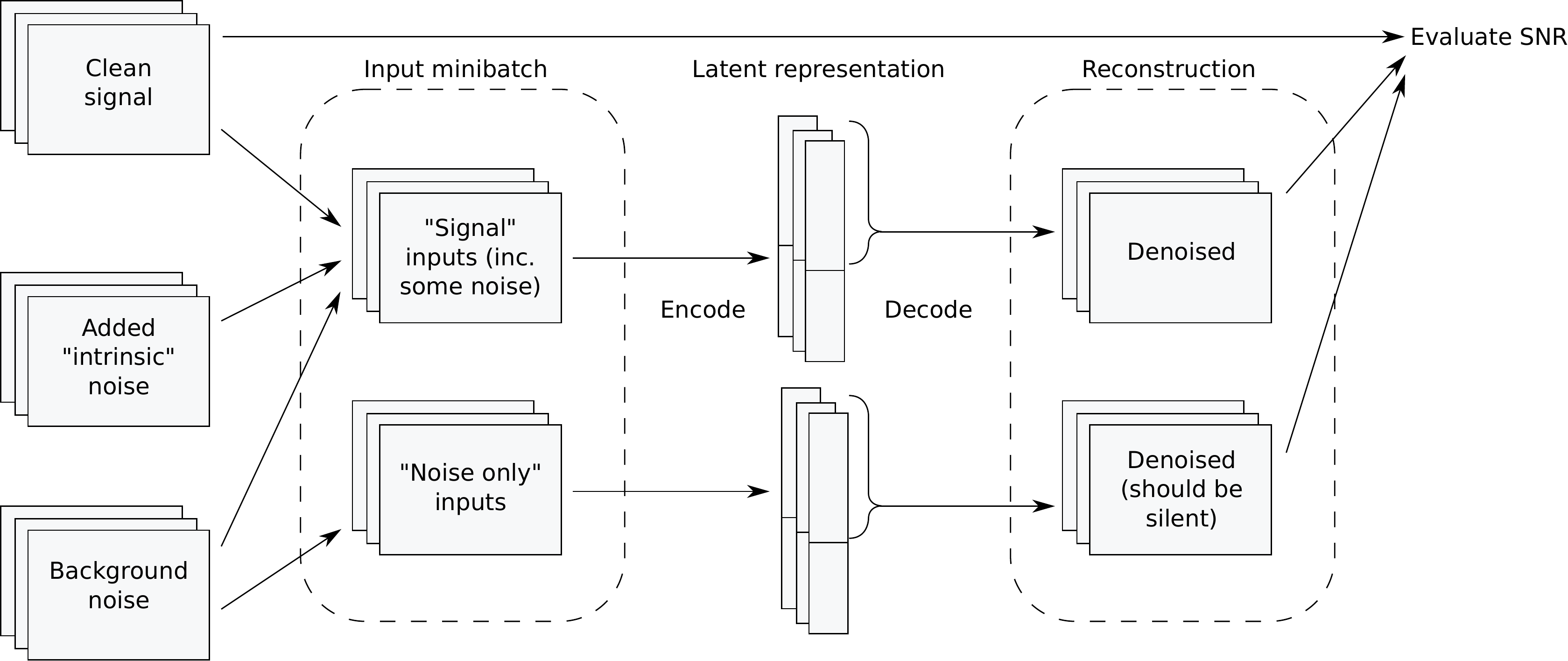}}
  \caption{For evaluation only (not for application), we construct our ``signal+noise'' examples from clean signal plus added intrinsic noise, so that we can evaluate SNR against the truly clean signal which the system never accesses.}
  \label{fig:partitionaesystem_denoi_eval}
\end{figure}

Our audio sources use sample rate 22.05 kHz,
analysed using 128 bin FFTs with 50\% hop,
and the frequency axis then reduced to 32 bins of interest for the birdsong (1.7--7.2 kHz).
Within this frequency band, we added intrinsic noise to give an SNR -10 dB, then extrinsic noise to give an SNR of -30 dB.
The extrinsic noise corresponds to the ``noise-only'' items that would be provided in practice.
The intrinsic noise is added for evaluation only, to judge whether the system can remove it (Figure \ref{fig:partitionaesystem_denoi_eval}).
We study two cases where the intrinsic and extrinsic noise sources are matched or unmatched.

\begin{figure}[t]
  \centering
  \centerline{\includegraphics[width=0.9\columnwidth,clip,trim=13mm 10mm 20mm 14mm]{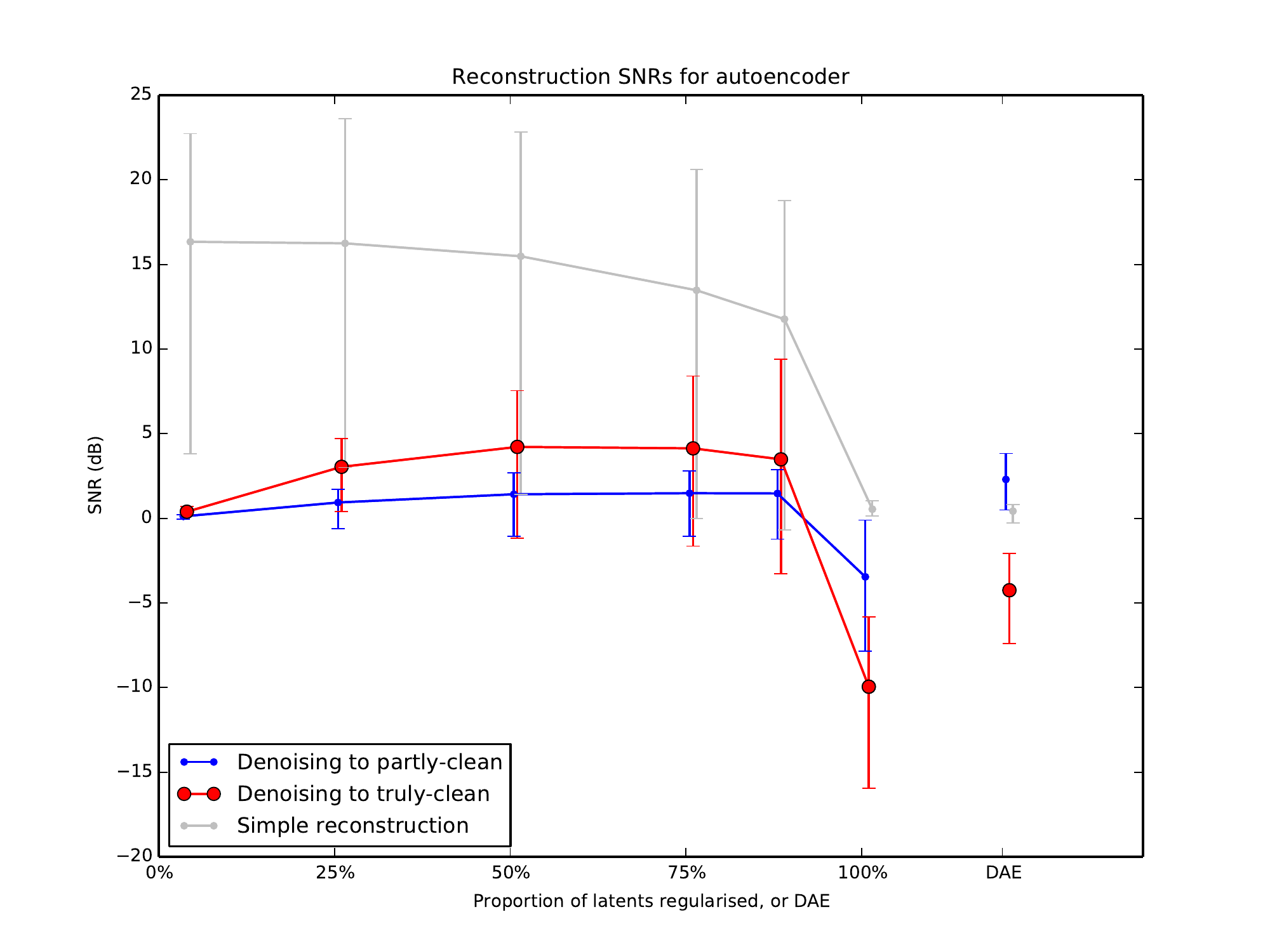}}
  \centerline{\includegraphics[width=0.9\columnwidth,clip,trim=13mm 5mm 20mm 14mm]{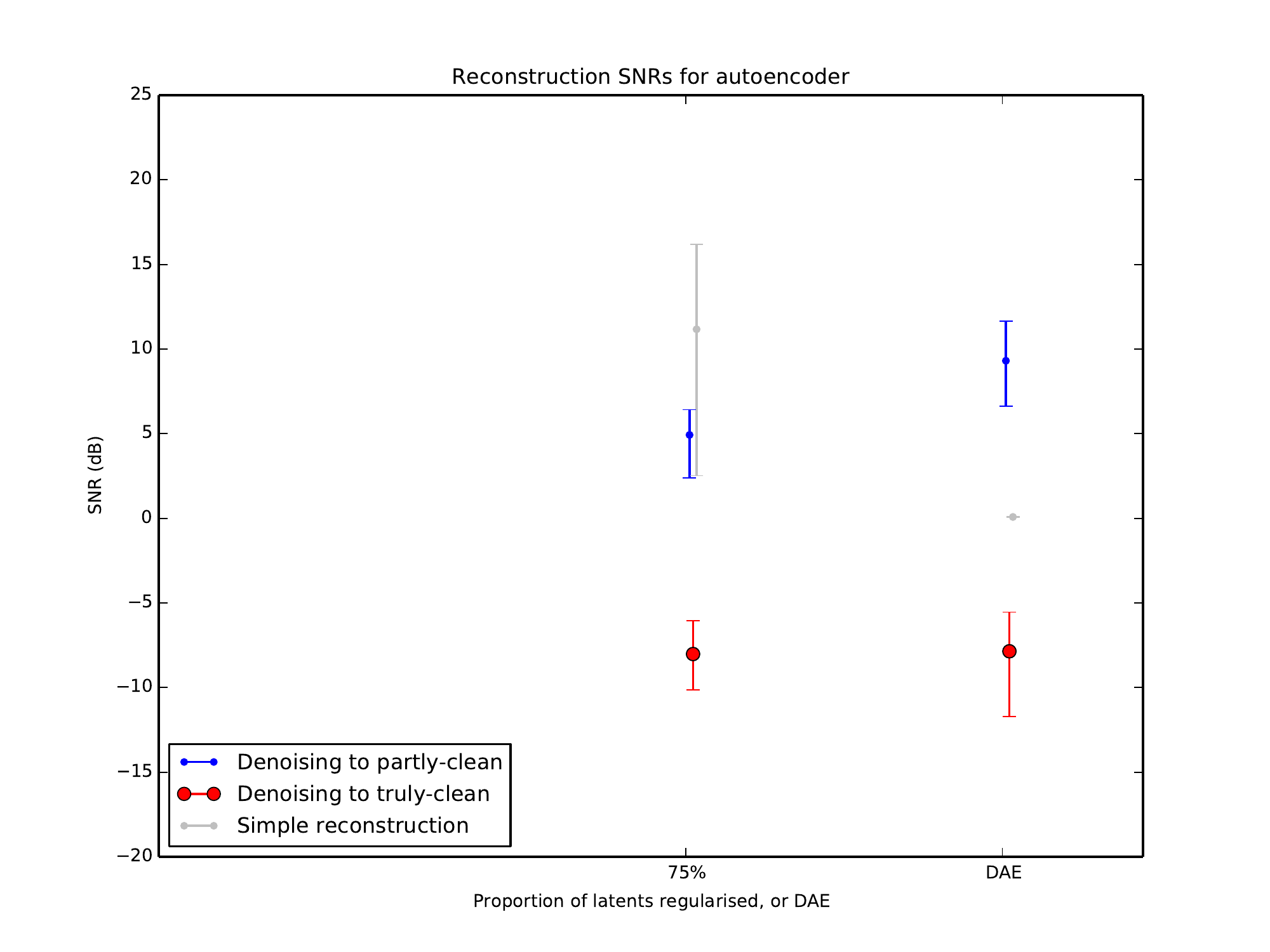}}
  \caption{Reconstruction SNRs measured on the validation data, for the partitioned AE or DAE. Intrinsic and extrinsic noise may be matched (upper plot) or unmatched (lower plot). Experiments are performed separately with three different noise source recordings, and the error bars show the range across three experiments. The large error bars are due to individual differences and not random variation (inspected by eye).}
  \label{fig:runplots}
\end{figure}

We test our partitioned AE at various settings for the proportion of latents regularised,
and also a standard DAE using the same training data, configured to use the extrinsic noise as the additive corruption $u(\cdot)$ in the DAE training process.
In all cases we fix $\lambda=0.75$ and train with minibatches of size 16, and $10^6$ iterations of SGD.

Results evaluated on separately-sampled validation data (Figure \ref{fig:runplots}) show a number of interesting properties.
Firstly, all methods perform best against their explicit objective in every case:
i.e.~their best SNR is the one relating to their objective function (simple reconstruction for the proposed system,
reconstruction of the partly-clean input for the DAE).
However, the statistic of interest is reconstruction of the truly-clean spectrogram.
In matched conditions (upper plot), the proposed system strongly outperforms the DAE on this measure.
Its strong performance is stable across a broad range of settings for the proportion of latents regularised
(all except the extremes).
The poor results at the setting with regularisation of all latents (100\%) confirm that the regularisation used is strong: if applied to all latents, it causes underfitting.
It is not merely the presence of regularisation that improves the results, but the partitioning scheme.

In unmatched conditions (lower plot) the strong performance is not sustained.
The DAE continues to perform well at its partial-denoising task,
but our proposed DAE fails to generalise across diverse background recordings.
It seems likely that this is due to the small size and depth of the current setup:
a single-layer autoencoder with only 32 convolutional filters has limited ability to approximate arbitrary functions.
The strong performance in matched conditions suggests further study of the method with deeper networks.
However, note that matched conditions are common in applications such as ours, where the noise-only examples can be taken from the main field recording sessions.

\section{Conclusions}
\label{sec:conc}

We have introduced a partitioned autoencoder that can learn to denoise data with better fidelity than a standard DAE,
in the common practical case where no noise-free examples are available for training.
In matched conditions this partitioned scheme makes better use of the available data than a DAE.
Unlike a DAE, our partitioned autoencoder learns to represent the signal and the noise content, and can reconstruct either or both.
This may be part of its advantage over a standard DAE, which does not learn to represent the noise content.
Further work will explore deeper/wider architectures to improve the generality, and use the representations to support classification and other tasks.

\section{Acknowledgments}
\label{sec:ack}

We gratefully thank Pavel Linhart for recording the chiff chaff bird sounds used in this study.

DS is supported by EPSRC Fellowship EP/L020505/1.

\clearpage

\bibliographystyle{IEEEbib}
\bibliography{../../../refs}

\end{document}